\documentclass[3p,times,11pt]{elsarticle}

\usepackage{amsmath,amssymb,amsfonts}
\usepackage{graphicx}
\usepackage{textcomp}
\usepackage{xcolor}
\usepackage{booktabs}
\usepackage{array}
\usepackage{url}
\usepackage{tikz}
\usepackage{pgfplots}
\pgfplotsset{compat=1.17}
\usepgfplotslibrary{fillbetween}
\usetikzlibrary{arrows.meta,positioning,fit,backgrounds,calc}

\journal{Applied Soft Computing}

\tikzset{
  block/.style={rectangle, rounded corners=2pt, draw=black!70, fill=#1,
                minimum height=7mm, inner sep=4pt, align=center, font=\scriptsize},
  stage/.style={block=blue!8},
  qr/.style={block=orange!14},
  cls/.style={block=teal!12},
  io/.style={rectangle, draw=black!55, fill=black!4, minimum height=7mm,
             inner sep=4pt, align=center, font=\scriptsize},
  ar/.style={-{Stealth[length=2mm]}, draw=black!70, line width=0.5pt},
  grp/.style={draw=black!30, dashed, rounded corners=3pt, inner sep=4pt},
}

\newcommand{\TQRNNd}{TQRNN$_{30\mathrm{d}}$}
\newcommand{\TQRNNh}{TQRNN$_{70\mathrm{h}}$}

\begin{document}

\begin{frontmatter}

\title{Quantile-Led Feature Extraction for Multi-Horizon Predictive Maintenance in Industrial Manufacturing Systems}

\author[herts]{David J. Poland\corref{cor1}}
\ead{d.j.poland@herts.ac.uk}
\author[herts]{Daniele Ravi}
\ead{d.ravi@herts.ac.uk}
\author[herts]{Na Helian}
\ead{n.helian@herts.ac.uk}

\cortext[cor1]{Corresponding author.}
\affiliation[herts]{organization={Department of Computer Science, University of Hertfordshire},
            addressline={College Lane},
            city={Hatfield},
            postcode={AL10 9AB},
            country={United Kingdom}}

\begin{abstract}
In data-driven predictive maintenance (PdM), feature extraction is usually
treated as fixed preprocessing: a descriptor set is chosen once and reused
unchanged while the downstream model is varied or the forecasting horizon is
extended. Across all four established extraction families, manual and domain-driven, classical data-driven, deep representation learning, and
hybrid multimodal designs, the representation is typically optimised for,
and evaluated at, a single prediction horizon, and systematic evidence on how
it should change as the horizon extends remains sparse. This paper isolates
the representation-learning stage and presents a quantile-led
feature-extraction framework in which a dual-stage MLP--QRNN
(quantile-regression neural network) hierarchy converts high-volume
multivariate sensor streams into compact, channel-resolved and
distribution-aware feature vectors under channel-resolved pinball-loss
supervision: QRNN$_1$ learns a broad ten-quantile conditional distribution
per channel, and a skip-connected QRNN$_2$ refines a retained mid-tail
quantile set. A fixed thirteen-pipeline staged ablation spans 1-hour,
70-hour, and 30-day regimes on an industrial deployment of 72 machines across
9 facilities, with the downstream temporal classifier held fixed. All else
held constant, increasing the retained mid-tail set from two to
four quantiles consistently improves
30- and 60-minute F1-score, reaching 75.92\% and 72.44\% with attention
enabled. The representation transfers beyond its design horizon only when
extractor capacity, temporal embedding, activation strategy, and sensor
breadth are scaled with it: the unmodified short-horizon extractor falls to
42.90\% F1 at 70 hours, whereas horizon-conditioned extractors reach 60.38\%
at 70 hours and 79.97\% at 30 days. Compressing roughly 180{,}000
reference-grid samples per channel each hour into $d_f\in\{86,172,324\}$
features, the framework provides deployment-scale evidence that PdM feature
extraction should be designed as a horizon-dependent representational stage
rather than as fixed preprocessing.
\end{abstract}


\end{frontmatter}


\section{Introduction}

Data-driven predictive maintenance (PdM) infers equipment
health from sensor evidence so that intervention can occur before failure,
replacing reactive correction and fixed-interval prevention with
condition-informed planning~\cite{Sellitto2023,Sharma2022,Abidi2022}. Modern
embedded sensing makes high-volume multivariate monitoring routine, but the
predictive value of that volume depends almost entirely on the representation
extracted from it. Incipient degradation in manufacturing machinery rarely
presents as a large point anomaly; it appears first as widening dispersion,
asymmetric tail growth, or small shifts in the relative position of
observations within the operating distribution~\cite{Chen2023,Mao2024}.
Point-estimate features obscure exactly these effects through averaging.

Most reported PdM systems nevertheless treat feature extraction as a fixed
preprocessing stage: a set of statistical, spectral, or learned descriptors is
chosen once and reused unchanged while the downstream model is varied or the
forecasting horizon is extended~\cite{Qiu2023,Serradilla2022,Solis2024}. This
paper takes the opposite view. It isolates the representation-learning stage
of an industrial fault-prediction pipeline and asks a single question: which
aspects of quantile-based feature extraction contribute to stability,
sensitivity, and scalability as the prediction horizon expands from one hour
to thirty days?

This paper presents a \emph{quantile-led} feature-extraction framework for
multi-horizon predictive maintenance, in which a dual-stage MLP--QRNN
hierarchy converting high-volume multivariate sensor streams into compact,
channel-resolved, distribution-aware feature vectors under channel-resolved
pinball-loss supervision. Concretely, a dual-stage quantile-regression neural
network (QRNN) hierarchy, supported by multilayer-perceptron (MLP) encoders,
converts each window of multivariate sensor behaviour into a compact,
channel-resolved, distribution-aware feature vector. The first stage, QRNN$_1$, estimates a broad ten-quantile conditional
distribution for every sensor channel; the second stage, QRNN$_2$, refines a
small retained mid-tail quantile set into the classifier-facing
representation. The hierarchy is deliberately separated from classifier design. The original short-horizon TQRNN formulation was introduced in earlier peer-reviewed work~\cite{Poland2024}, where quantile-derived representations were combined with Transformer-based temporal classification for near-term industrial machine-health prediction. The present study extends the feature-extraction methodology beyond that original short-horizon formulation and evaluates how the representation must change across intermediate- and long-horizon regimes. Within each experimental regime, the downstream temporal decision stage is held fixed so that the comparisons reported here isolate changes in the upstream feature representation rather than changes in classifier design.

The framework is evaluated on a multi-site industrial deployment of 72
machines across 9 facilities through a fixed, staged thirteen-pipeline
ablation programme spanning three forecasting regimes:
the short-horizon TQRNN regime (10\,s--60\,min, 43 sensor channels), the
extended-hour \TQRNNh{} regime (36--70\,h), and the day-scale \TQRNNd{} regime
(7--30 days, 81 sensor channels). These correspond, respectively, to the
short-, medium-, and long-horizon operational regimes defined in the
companion long-horizon study. The strongest configuration of each stage
is carried forward unchanged as the representational baseline of the next, so
that the cost of transferring a representation beyond its design horizon is
quantified directly.

The main contributions are:
\begin{enumerate}
\item A dual-stage MLP--QRNN feature-extraction hierarchy that couples
broad-distribution learning with targeted mid-tail refinement under
channel-resolved pinball-loss supervision, producing compact quantile-state
feature vectors from raw multi-rate industrial sensor data.
\item A controlled demonstration that increasing the retained mid-tail
quantile set from two to four levels
($\alpha\in\{0.25,0.40,0.60,0.75\}$) improves predictive robustness as
temporal separation grows, with all other components held fixed.
\item A horizon-conditioned scaling principle, covering feature capacity,
temporal embedding, activation strategy, and sensor breadth, under which the
four-quantile representation transfers from the 1-hour to the 70-hour and
30-day regimes.

\end{enumerate}

The remainder of the paper is organised as follows. Section~\ref{sec:related} reviews
related work. Section~\ref{sec:data} describes the industrial data foundation and
temporal representation. Section~\ref{sec:method} specifies the quantile-led
feature-extraction hierarchy and its horizon-conditioned variants. Section~\ref{sec:setup}
defines the staged experimental design, and Section~\ref{sec:results} reports and discusses
the results. Section~\ref{sec:conclusion} concludes.

\section{Related Work}
\label{sec:related}

\subsection{Feature Extraction for Industrial Time Series}

Feature extraction methods for industrial prognostics span four broad
families: (i) manual and domain-driven extraction, (ii) classical data-driven
extraction, (iii) deep representation learning, and (iv) hybrid and
multimodal designs. Each is outlined below.

Manual and domain-driven extraction encodes engineering knowledge
through statistical moments, spectral bands, envelope analysis, and kurtosis-
or deconvolution-based indicators. Descriptors of this kind remain effective
wherever the failure physics of the monitored asset are well
characterised~\cite{Solis2024,Hashim2023}. Classical data-driven
extraction applies generic transforms and shallow learners---principal
component analysis, wavelet decompositions, and clustering-based
descriptors---to derive a compact and discriminative summary of the raw
signal. Dimensionality reduction is the mechanism by which such a summary is
obtained rather than the objective of feature extraction itself; the
objective is a representation that retains the information relevant to the
prognostic task while discarding the rest. These methods achieve that
reduction with shallow, computationally inexpensive models, without recourse
to deep neural architectures~\cite{Kim2022}. Deep representation
learning replaces hand-crafted descriptors with hierarchies learned end to
end, using convolutional, recurrent, and autoencoding networks to capture
non-linear cross-channel and temporal structure~\cite{Zhang2022,Fu2023,Lai2019}.
Hybrid and multimodal designs combine these elements, frequently fusing
heterogeneous sensor groups with metadata or operational
context~\cite{Tsanousa2022,Gawde2024}. Across all four families, however, the
extracted representation is typically optimised for, and evaluated at, a
single prediction horizon; systematic evidence on how a representation should
change as the horizon extends remains sparse~\cite{Fernandes2022,Qiu2023}.

\subsection{Quantile Regression for Uncertainty-Aware Features}

Quantile regression estimates conditional quantiles rather than a single
expected value, and is therefore informative when the shape of the predictive
distribution matters~\cite{Koenker1978,Cannon2018}. In prognostics it has been
used in two ways. The first is direct probabilistic forecasting, in which
multiple quantiles are predicted jointly to give uncertainty-calibrated
degradation envelopes and risk bands~\cite{Jahangir2023,Zhang2024Q}. The
second, adopted here, is distributional feature extraction: quantiles
summarise the spread and asymmetry of a signal window before a downstream
prognostic or classification step~\cite{Mao2024,Dong2025}. The shared
motivation is that tail behaviour and inter-quantile spread change before the
conditional mean does. Because independently estimated quantiles can cross,
monotonicity-preserving formulations are applied so that the learned envelope
remains a coherent distributional representation~\cite{Cannon2018}. Prior
quantile-based prognostic work has largely used a single quantile stage and a
fixed quantile set; the cascaded broad-to-refined hierarchy and the explicit
study of retained-quantile cardinality across horizons developed in this paper
have not, to the authors' knowledge, been reported.

\subsection{Temporal Modelling over Learned Features}

Transformer-based models capture long-range dependencies in industrial
sequences through attention~\cite{Vaswani2017,Wen2023}, and have been applied
to anomaly detection, remaining-useful-life estimation, and multi-horizon
forecasting~\cite{LiJ2024,Bampoula2024,Lim2021}. Full attention over raw
high-rate streams is computationally demanding and can dilute temporal
ordering at very long sequence lengths~\cite{Grigsby2021}. This cost grows
quadratically with sequence length, which motivates hierarchical designs in
which attention operates over compressed temporal summaries rather than raw
samples~\cite{LiJ2024}. In this paper the
Transformer-based decision stage is exactly such a consumer: it operates on
hourly quantile-state feature vectors produced by the proposed extractor, so
that the effective sequence length is the number of hourly words rather than
the number of raw measurements. The decision stage itself is held fixed within each experimental regime.
At the longest horizon, the extracted representation is supplied to the
multi-stream temporal-fusion decision stage used in the deployment-scale
pipeline; the present paper evaluates the upstream representation rather than
claiming the downstream classifier as a feature-extraction contribution.

\section{Industrial Data Foundation and Temporal Representation}
\label{sec:data}

\subsection{Deployment Context}

The evaluation uses data from a manufacturing deployment comprising 72
machines across 9 facilities in EMEA and North America. The machines belong
to the same broad high-speed manufacturing asset and product family, use consistent sensor placement, supporting a controlled multi-site evaluation
while still exposing the extractor to variation in facility conditions,
materials, operation, and maintenance practice. In this environment unplanned
downtime creates economic and logistical pressure for early fault detection,
and failure modes can cascade rapidly, so representations must be both
sensitive to incipient distributional change and computationally efficient
for continuous use.

\subsection{Sensor Configuration}

Each machine is instrumented with heterogeneous 12-bit sensors producing
integer readings in $[0,4095]$ across vibration/motion, fluid/flow, thermal,
pressure, tooling/actuation, and high-frequency acceleration groups, sampled
at native rates between 0.033\,Hz and 50\,Hz. The short-horizon and
extended-hour regimes use a 43-channel configuration ($C_s=43$); the
day-scale regime expands this to 81 channels ($C_s=81$) to provide the
sensor breadth required for slow-evolving degradation modes.

\subsection{Hourly Words and Documents}

The native multi-rate streams are aligned onto a common 20\,ms causal
reference grid, with lower-frequency channels forward-aligned so that no
future information enters any feature. One hour of aligned behaviour, of the
order of $N_h=180{,}000$ reference-grid steps per channel, forms one
\emph{hourly word} $w_k$, and an ordered sequence of $K$ words forms the
input document for a given horizon (e.g., $K=720$ for the 30-day document).
The feature extractor studied in this paper is the mapping
$\mathbf{f}_k=\Phi_Q(w_k)\in\mathbb{R}^{d_f}$ from each hourly word to a
compact quantile-state vector; the downstream classifier then reasons over
the document $[\mathbf{f}_1,\ldots,\mathbf{f}_K]$.

For horizons beyond one hour, the physical sensor vector is augmented with an
18-dimensional temporal embedding encoding cyclical and contextual
information (time of day, shift, and schedule structure), raising the
effective extractor input from 43 to 61 dimensions in the \TQRNNh{} regime
and from 81 to 99 dimensions in the \TQRNNd{} regime.

\subsection{Labels and Splits}

Each hourly word is labelled normal or abnormal using maintenance records,
operator logs, PLC fault reports, and expert validation. The short-horizon
regime, constructed during the initial single-site pilot, uses a chronological
60/20/20 train/validation/test split; non-overlapping hourly documents make
additional purge intervals unnecessary. The extended-hour and day-scale regimes
are built across the full multi-site deployment and use a machine-disjoint split
(43 training, 14 validation, and 15 test machines), so that no machine
contributes to more than one partition and reported transfer is free of
machine-level leakage. All configurations within a regime share identical partitions,
horizons, and labels.

\section{Quantile-Led Feature Extraction}
\label{sec:method}

\subsection{Channel-Resolved Quantile Objective}

For sensor channel $i$ at temporal position $t$, the observed 12-bit target is
$y_{ti}\in[0,4095]$, and the multichannel observation is
$\mathbf{x}_t=[x_{t1},\ldots,x_{tC_s}]\in\mathbb{R}^{C_s}$. A predicted
conditional quantile $\hat{q}_\alpha$ at level $\alpha\in(0,1)$ is trained
with the pinball loss
\begin{equation}
\mathcal{L}_{\alpha}(y,\hat{q}_\alpha)
=
\alpha\max(y-\hat{q}_\alpha,0)
+
(1-\alpha)\max(\hat{q}_\alpha-y,0),
\label{eq:pinball}
\end{equation}
which penalises over- and under-estimation asymmetrically. For a quantile set
$\mathcal{A}$ and mini-batch $\mathcal{B}$, the aggregate stage objective is
\begin{equation}
\mathcal{L}^{\mathrm{QRNN}}
=
\frac{1}{|\mathcal{B}|\,|\mathcal{A}|\,C_s}
\sum_{b\in\mathcal{B}}
\sum_{\alpha\in\mathcal{A}}
\sum_{i=1}^{C_s}
\mathcal{L}_{\alpha}\!\left(
y_{b,i},\,\hat{q}^{(i)}_{\alpha}(\mathbf{x}_b)\right).
\label{eq:qrnnloss}
\end{equation}
Although the intermediate MLP embeddings mix information across channels, the
supervised outputs remain \emph{channel-resolved}: each stage predicts a
separate quantile estimate per channel and per level, and the loss is
evaluated per channel--quantile pair before averaging. Global cross-channel
context conditions every prediction, but the quantile errors remain
sensor-specific. To prevent quantile crossing, predicted quantiles are
ordered following non-crossing quantile-regression
principles~\cite{Cannon2018} before downstream feature construction.

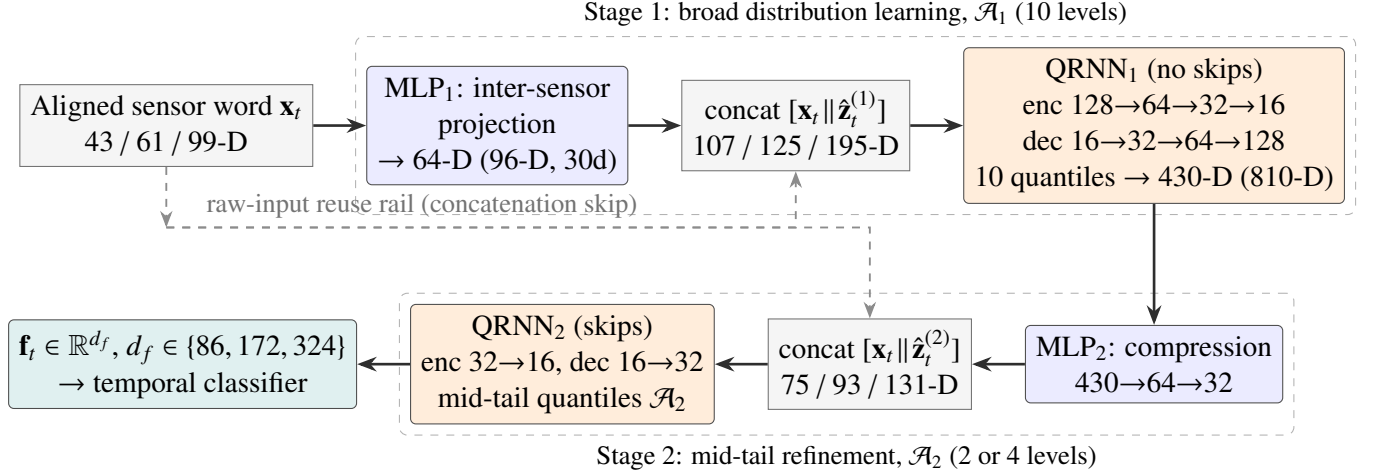
\begin{figure}[!t]
\centering
\begin{tikzpicture}[node distance=6mm and 7mm,
  io/.append style={font=\normalsize},
  stage/.append style={font=\normalsize},
  qr/.append style={font=\normalsize},
  cls/.append style={font=\normalsize},
  trunk/.style={-{Stealth[length=2.6mm]}, draw=black!82, line width=1.0pt},
  skip/.style={-{Stealth[length=2.2mm]}, draw=black!45, line width=0.7pt, dashed}]
\node[io] (inp) {Aligned sensor word $\mathbf{x}_t$\\43 / 61 / 99-D};
\node[stage, right=of inp] (mlp1) {MLP$_1$: inter-sensor\\projection\\$\to$ 64-D (96-D, 30d)};
\node[io, right=of mlp1] (cat1) {concat $[\mathbf{x}_t\!\parallel\!\hat{\mathbf{z}}^{(1)}_t]$\\107 / 125 / 195-D};
\node[qr, right=of cat1] (q1) {QRNN$_1$ (no skips)\\enc $128{\to}64{\to}32{\to}16$\\dec $16{\to}32{\to}64{\to}128$\\10 quantiles $\to$ 430-D (810-D)};
\node[stage, below=16mm of q1] (mlp2) {MLP$_2$: compression\\$430{\to}64{\to}32$};
\node[io, left=of mlp2] (cat2) {concat $[\mathbf{x}_t\!\parallel\!\hat{\mathbf{z}}^{(2)}_t]$\\75 / 93 / 131-D};
\node[qr, left=of cat2] (q2) {QRNN$_2$ (skips)\\enc $32{\to}16$, dec $16{\to}32$\\mid-tail quantiles $\mathcal{A}_2$};
\node[cls, left=of q2] (out) {$\mathbf{f}_t\in\mathbb{R}^{d_f}$, $d_f\in\{86,172,324\}$\\$\to$ temporal classifier};
\draw[trunk] (inp) -- (mlp1);
\draw[trunk] (mlp1) -- (cat1);
\draw[trunk] (cat1) -- (q1);
\draw[trunk] (q1) -- (mlp2);
\draw[trunk] (mlp2) -- (cat2);
\draw[trunk] (cat2) -- (q2);
\draw[trunk] (q2) -- (out);
\coordinate (rail) at ($(inp.south)+(0,-8mm)$);
\draw[skip] (inp.south) -- (rail);
\draw[skip] (rail) -| (cat1.south);
\draw[skip] (rail) -| (cat2.north);
\node[font=\small, text=black!55, anchor=west] at ($(rail)+(4mm,2.8mm)$)
  {raw-input reuse rail (concatenation skip)};
\begin{scope}[on background layer]
\node[grp, fit=(mlp1)(q1), label={[font=\small]above:Stage 1: broad distribution learning, $\mathcal{A}_1$ (10 levels)}] {};
\node[grp, fit=(q2)(mlp2), label={[font=\small]below:Stage 2: mid-tail refinement, $\mathcal{A}_2$ (2 or 4 levels)}] {};
\end{scope}
\end{tikzpicture}
\caption{Dual-stage quantile-led feature extractor. MLP$_1$ encodes
inter-sensor structure; QRNN$_1$ learns a broad ten-quantile conditional
distribution per channel through a deeper encoder--decoder without skip
connections; MLP$_2$ compresses the 430-dimensional (810-dimensional in the
81-channel regime) broad representation; QRNN$_2$ refines the retained
mid-tail quantiles through a shallow skip-connected encoder--decoder. The
output $\mathbf{f}_t$ is the channel-resolved quantile-state feature vector
consumed by the downstream temporal classifier. Dimensions in parentheses
denote the day-scale 81-channel configuration; slash-separated dimensions
list the short-horizon, extended-hour, and day-scale regimes in order.}
\label{fig:extractor}
\end{figure}

\subsection{Dual-Stage Hierarchy}

The extractor separates \emph{broad distribution learning} from
\emph{targeted mid-tail refinement} (Fig.~\ref{fig:extractor}). The
short-horizon 43-channel specification is given first; horizon-conditioned
variants follow in Section~\ref{sec:scaling}.

\subsubsection{MLP$_1$: inter-sensor projection}
A front-end MLP captures non-linear cross-channel structure:
\begin{align}
\mathbf{h}^{(1)}_t&=\operatorname{LReLU}\!\left(W^{(1)}\mathbf{x}_t+\mathbf{b}^{(1)}\right),\\
\mathbf{h}^{(2)}_t&=\operatorname{LReLU}\!\left(W^{(2)}\mathbf{h}^{(1)}_t+\mathbf{b}^{(2)}\right),\\
\hat{\mathbf{z}}^{(1)}_t&=W^{(\mathrm{out})}\mathbf{h}^{(2)}_t+\mathbf{b}^{(\mathrm{out})},
\end{align}
with layer widths $43\!\to\!128\!\to\!64\!\to\!64$ and Leaky ReLU
($\operatorname{LReLU}$) activations. The 64-dimensional embedding is
concatenated with the raw input to form the first-stage QRNN input
$\mathbf{x}^{(1)}_t=[\mathbf{x}_t\parallel\hat{\mathbf{z}}^{(1)}_t]\in\mathbb{R}^{107}$,
so that distributional learning is conditioned on both measured and learned
cross-channel context.

\subsubsection{QRNN$_1$: broad multi-quantile prediction}
QRNN$_1$ uses a deeper encoder--decoder \emph{without} skip connections to
support stable global distribution learning. The encoder compresses
$107\!\to\!128\!\to\!64\!\to\!32\!\to\!16$ and the decoder expands
$16\!\to\!32\!\to\!64\!\to\!128$, each layer applying
$\operatorname{LReLU}(W\mathbf{h}+\mathbf{b})$. A final projection maps the
decoder state to the channel-resolved multi-quantile output over the broad
set
\begin{equation}
\mathcal{A}_1=\{0.01,\!0.10,\!0.20,\!0.25,\!0.50,\!0.60,\!0.75,\!0.80,\!0.90,\!0.99\},
\end{equation}
giving
$\hat{\mathbf{Q}}^{(1)}_t\in\mathbb{R}^{43\times10}$, i.e.
$\operatorname{vec}(\hat{\mathbf{Q}}^{(1)}_t)\in\mathbb{R}^{430}$. Covering
lower-tail, central, and upper-tail levels lets QRNN$_1$ learn the overall
shape of each channel's conditional distribution under normal and abnormal
operation.

\subsubsection{MLP$_2$: mid-tail compression}
MLP$_2$ condenses the broad representation through
$430\!\to\!64\!\to\!32$ with Leaky ReLU activations, producing
$\hat{\mathbf{z}}^{(2)}_t\in\mathbb{R}^{32}$. This learned latent space
emphasises the mid-tail region, where early anomaly signatures are expected
to be most visible, and filters less critical distributional detail. The
compressed embedding is re-anchored to the measured input,
$\mathbf{x}^{(2)}_t=[\mathbf{x}_t\parallel\hat{\mathbf{z}}^{(2)}_t]\in\mathbb{R}^{75}$.

\subsubsection{QRNN$_2$: skip-connected mid-tail refinement}
The second stage uses a shallow encoder--decoder \emph{with} skip
connections, e.g.
\begin{equation}
\mathbf{h}'^{(1)}
=\operatorname{LReLU}\!\left(W'^{(1)}\mathbf{x}^{(2)}_t+\mathbf{b}'^{(1)}\right)
+S^{(1)}\mathbf{x}^{(2)}_t,
\end{equation}
with encoder $75\!\to\!32\!\to\!16$ and decoder $16\!\to\!16\!\to\!32$, where
$S^{(\cdot)}$ and the decoder skip matrices $R^{(\cdot)}$ preserve coarse
context and support gradient flow while enabling fine-grained quantile
correction. The output projection yields the retained mid-tail prediction
$\hat{\mathbf{Q}}^{(2)}_t\in\mathbb{R}^{43\times|\mathcal{A}_2|}$, and the
classifier-facing feature vector is
\begin{equation}
\mathbf{f}_t=\operatorname{vec}\bigl(\hat{\mathbf{Q}}^{(2)}_t\bigr)
\in\mathbb{R}^{43\,|\mathcal{A}_2|}.
\end{equation}
In the baseline two-quantile configuration,
$\mathcal{A}_2=\{0.25,0.75\}$ and $\mathbf{f}_t\in\mathbb{R}^{86}$. The
asymmetric strategy, deep without skips for global distribution learning,
shallow with skips for refinement, balances distributional coverage, anomaly
sensitivity, and latency.

\subsection{Two-Tier Training Procedure}

QRNN$_1$ is trained first by minimising \eqref{eq:qrnnloss} over
$\mathcal{A}_1$ until multi-quantile coverage is calibrated across channels
($43\times10=430$ channel--quantile loss terms per example). Its parameters
are then frozen, and MLP$_2$ and QRNN$_2$ are trained to refine the retained
mid-tail structure by minimising \eqref{eq:qrnnloss} over $\mathcal{A}_2$.
This staging first captures broad distributional structure and then
concentrates capacity where pre-failure distributional widening is most
relevant to the downstream classifier. Optimisation uses Adam with
learning-rate decay, dropout, and validation-based early stopping.

\subsection{The Four-Quantile Mid-Tail Variant}

A closely related variant retains the identical
MLP$_1\!\to$QRNN$_1\!\to$MLP$_2\!\to$QRNN$_2$ pipeline and training
procedure but increases the refined set from two to four mid- and near-tail
levels,
\begin{equation}
\mathcal{A}_2=\{0.25,0.40,0.60,0.75\},
\end{equation}
raising the output dimensionality from $43\times2=86$ to $43\times4=172$
(Pipeline P5 in the ablation). Because every other component is held fixed,
any performance difference is attributable to the resolution of the refined
mid-tail representation. The hypothesis under test is that denser mid-tail
sampling improves sensitivity to gradual distributional drift, with the gain
growing as the prediction horizon increases.

\subsection{Horizon-Conditioned Scaling}
\label{sec:scaling}

The same dual-stage hierarchy is preserved as the framework extends from the
1-hour regime to \TQRNNh{} and \TQRNNd{}; the scaling principle is not to
replace the extractor but to increase its representational capacity in line
with the horizon (Table~\ref{tab:regimes}). Three coordinated extensions are
applied. First, \emph{temporal context}: the 18-dimensional temporal
embedding augments the physical input (61-D at 70\,h; 99-D at 30 days) so
that periodic structure and slow-evolving patterns are visible to the
extractor. Second, \emph{capacity}: MLP encoders are widened and QRNN stacks
deepened; in the day-scale configuration MLP$_1$ outputs a 96-dimensional
embedding (versus 64-D at shorter horizons) and QRNN$_1$ produces
$81\times10=810$ broad quantile outputs. Third, \emph{activation strategy}:
Pipelines P1--P8 use Leaky ReLU, whereas the integrated extended-hour
configuration (P9) and all day-scale pipelines (P10--P13) adopt PReLU, whose
learnable negative slope $a_{\mathrm{PReLU}}$ supports long-horizon feature
learning on heavy-tailed inputs. Finally, \emph{sensor breadth} expands from
43 to 81 channels at day scale, so the four-quantile representation grows to
$81\times4=324$ dimensions. Across regimes the extractor therefore
compresses approximately $180{,}000$ reference-grid samples per channel each
hour into $d_f\in\{86,172,324\}$ features per hourly word.

\begin{table}[!t]
\centering
\caption{Horizon-Conditioned Extractor Configurations}
\label{tab:regimes}
\setlength{\tabcolsep}{2.6pt}
\footnotesize
\begin{tabular}{lccc}
\toprule
 & TQRNN & \TQRNNh & \TQRNNd \\
\midrule
Horizons & 10\,s--60\,min & 36--70\,h & 7--30 day \\
Physical channels $C_s$ & 43 & 43 & 81 \\
Temporal embedding & -- & 18-D & 18-D \\
Extractor input & 43-D & 61-D & 99-D \\
MLP$_1$ output & 64-D & 64-D & 96-D \\
QRNN$_1$ output & 430-D & 430-D & 810-D \\
Retained quantiles $|\mathcal{A}_2|$ & 2 / 4 & 4 & 4 \\
Feature dim.\ $d_f$ & 86 / 172 & 172 & 324 \\
Activation & LReLU & LReLU/PReLU & PReLU \\
Pipelines & P1--P5 & P5--P9 & P9--P13 \\
\bottomrule
\end{tabular}
\end{table}

\section{Experimental Design}
\label{sec:setup}

\subsection{Staged Ablation Protocol}

\begin{figure}[!t]
\centering
\begin{tikzpicture}[node distance=3.5mm]
\node[stage, text width=62mm] (s1) {\textbf{Short horizon (TQRNN, 43 ch)}\\
P1 $QR^{1}$ \;|\; P2 $QR^{2}$ \;|\; P3 Transformer-only\\
P4 $All^{(QR^{2}_{2Q})}$ \;|\; \textbf{P5 $All^{(QR^{2}_{4Q})}$}};
\node[stage, below=of s1, text width=62mm] (s2) {\textbf{Extended hour (\TQRNNh, 61-D input)}\\
P5 (fixed ref.) \;|\; P6 $QR^{1}_{70h}$ \;|\; P7 $QR^{2}_{70h}$\\
P8 Transformer$_{70h}$ \;|\; \textbf{P9 $All_{70h}$}};
\node[stage, below=of s2, text width=62mm] (s3) {\textbf{Day scale (\TQRNNd, 81 ch, 99-D input)}\\
P9 (fixed ref.) \;|\; P10 $QR^{1}_{30d}$ \;|\; P11 $QR^{2}_{30d}$\\
P12 Transformer$_{30d}$ \;|\; \textbf{P13 $All_{30d}$}};
\draw[ar] (s1) -- node[right,font=\tiny]{P5 carried forward: 4Q mid-tail set} (s2);
\draw[ar] (s2) -- node[right,font=\tiny]{P9 carried forward: capacity + PReLU} (s3);
\end{tikzpicture}
\caption{Staged thirteen-pipeline ablation (P1--P13). The strongest configuration of
each regime (bold) is carried forward unchanged as the representational
baseline of the next, so transfer degradation is measured directly.}
\label{fig:roadmap}
\end{figure}
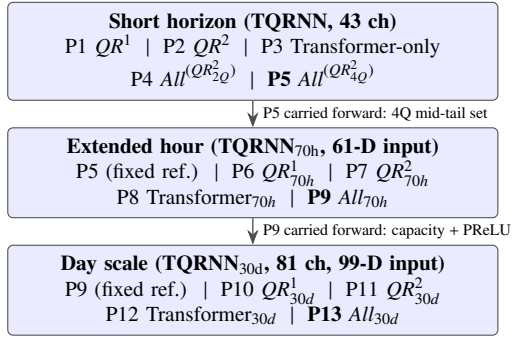

The evaluation is organised as a fixed staged ablation across thirteen
pipelines, P1--P13 (Fig.~\ref{fig:roadmap}). Each stage corresponds to a progressively
extended regime, and the strongest configuration of one stage is carried
forward \emph{unchanged} as the representational baseline of the next. This
design attributes observed gains to staged architectural extension rather
than uncontrolled re-optimisation, and directly quantifies the degradation
incurred when a representation is applied beyond its design horizon.

Each regime contrasts the same four configuration types, so that the effect
of hierarchical integration can be read consistently across horizons. The
\emph{broad-quantile stage alone} ($QR^{1}$) evaluates QRNN$_1$ in isolation,
without mid-tail refinement. The \emph{mid-tail refinement stage alone}
($QR^{2}$) evaluates QRNN$_2$ in isolation, without broad distributional
context. The \emph{Transformer-only} configuration operates directly on the
cleaned sensor input with no quantile extraction at all, and isolates the
contribution of temporal modelling. The \emph{integrated extractor} ($All$)
is the complete MLP$_1\!\to$QRNN$_1\!\to$MLP$_2\!\to$QRNN$_2$ hierarchy of
Section~\ref{sec:method}. Instantiating these four types across the three regimes yields the
thirteen pipelines P1--P13 that label the tables and figures throughout. In a
pipeline name, a subscript ($70h$, $30d$) marks the regime the configuration
was trained for, and the superscript on $All$ gives the retained mid-tail set
($2Q$ or $4Q$).

\emph{Short horizon (P1--P5).} The four types are evaluated on the 43-channel
input as P1 ($QR^{1}$), P2 ($QR^{2}$), and P3 (Transformer-only). The
integrated extractor is instantiated twice, as P4 ($All^{(QR^{2}_{2Q})}$) and
P5 ($All^{(QR^{2}_{4Q})}$), so that retained-quantile cardinality is varied
with everything else held fixed.

\emph{Extended hour (P5--P9).} The strongest short-horizon configuration, P5
with additive attention, is retained unchanged as a transferred reference.
Against it, the broad-quantile and mid-tail refinement stages are re-trained
under extended-hour conditions and evaluated in isolation as P6
($QR^{1}_{70h}$) and P7 ($QR^{2}_{70h}$), alongside a Transformer-only
variant, P8 (Transformer$_{70h}$), and the fully integrated extended-hour
extractor, P9 ($All_{70h}$).

\emph{Day scale (P9--P13).} P9 is retained unchanged as the transferred
bridge configuration. The broad-quantile and mid-tail refinement stages are
re-trained on the 81-channel input and evaluated in isolation as P10
($QR^{1}_{30d}$) and P11 ($QR^{2}_{30d}$), again alongside a day-scale
Transformer-only variant, P12 (Transformer$_{30d}$), and the fully integrated
day-scale extractor, P13 ($All_{30d}$).

\subsection{Decision Stage, Baselines, and Metrics}

Within each regime the downstream temporal decision model is held fixed so
that differences reflect the representation. Quantile features are consumed by a Transformer-based temporal decision stage
operating at word level. Within each forecasting regime, that decision stage
is held fixed while the upstream representation is varied. At the longest
horizon, the representation is supplied to the multi-stream temporal-fusion
decision stage used in the deployment-scale configuration. For external comparison in the short-horizon regime,
six classical and deep baselines, SVR, KNN, LSRN, LS-SVM, LSTM, and a
stand-alone Transformer, are evaluated using the same data representation as
the integrated two-quantile extractor, ensuring that differences primarily
reflect modelling capacity
rather than preprocessing. Performance is reported as F1-score, recall,
precision, and accuracy at fixed operating thresholds, with matched
train/validation/test partitions and horizons throughout. All models are
implemented in Python using PyTorch, NumPy, pandas, SciPy, and scikit-learn.

\section{Results and Discussion}
\label{sec:results}

\subsection{Short-Horizon Ablation (P1--P5)}

\begin{table}[!t]
\centering
\caption{Short-Horizon Feature-Extraction Ablation (Pipelines P1--P5). The Final Row Reports P5 with Additive Attention Enabled.}
\label{tab:short}
\setlength{\tabcolsep}{4.4pt}
\footnotesize
\begin{tabular}{l|cccc|cccc|cccc}
\toprule
& \multicolumn{4}{c|}{10 seconds} & \multicolumn{4}{c|}{30 minutes} & \multicolumn{4}{c}{60 minutes} \\
\cmidrule(lr){2-5}\cmidrule(lr){6-9}\cmidrule(l){10-13}
Pipeline & F1 & Rec & Prec & Acc\% & F1 & Rec & Prec & Acc\% & F1 & Rec & Prec & Acc\% \\
\midrule
P1: $QR^{1}$ & 84.75 & 83.30 & 88.00 & 83.74 & 50.54 & 48.51 & 52.76 & 51.65 & 51.76 & 50.19 & 53.44 & 54.27 \\
P2: $QR^{2}$ & 95.47 & 95.03 & 95.38 & 96.15 & 60.23 & 59.57 & 60.91 & 59.94 & 57.04 & 56.27 & 57.84 & 58.45 \\
P3: Transformer-only & 97.25 & 97.07 & 97.49 & 97.25 & 70.65 & 68.91 & 71.49 & 72.17 & 63.23 & 62.60 & 63.89 & 66.71 \\
P4: $All^{(QR^{2}_{2Q})}$ & 97.54 & 97.48 & 97.44 & 97.52 & 72.73 & 73.32 & 73.76 & 72.56 & 67.68 & 67.01 & 68.38 & 68.67 \\
P5: $All^{(QR^{2}_{4Q})}$ & 97.75 & 97.79 & 97.66 & 97.67 & 73.29 & 72.54 & 73.10 & 73.92 & 69.67 & 70.43 & 69.88 & 70.07 \\
P5 + additive attention & \textbf{98.39} & \textbf{98.58} & \textbf{97.92} & \textbf{97.90}
    & \textbf{75.92} & \textbf{75.77} & \textbf{74.33} & \textbf{74.15}
    & \textbf{72.44} & \textbf{72.20} & \textbf{73.65} & \textbf{73.84} \\
\bottomrule
\end{tabular}
\end{table}

Table~\ref{tab:short} reports the short-horizon ablation over 10-second,
30-minute, and 60-minute windows. The broad-quantile stage used alone (P1) performs adequately at 10 seconds but degrades sharply with horizon,
indicating that coarse multi-quantile coverage alone is insufficient. The
mid-tail refinement stage used alone (P2) achieves near-saturated 10-second
performance, reflecting the value of mid-tail calibration for rapid anomaly
detection, but also deteriorates at longer windows because it lacks broader
distributional context. The Transformer-only baseline (P3) is stronger at 30 and 60 minutes, confirming the value of direct temporal modelling, yet is
consistently outperformed by the quantile-integrated configurations.

The fully integrated configurations are strongest overall. With every other
component held fixed, extending the refinement stage from two mid-tail quantiles (P4) to four (P5) leaves the saturated 10-second result essentially unchanged
(97.54\% vs 97.75\% F1) and yields modest but consistent gains at 30 minutes
(72.73\%\,$\to$\,73.29\%) and 60 minutes (67.68\%\,$\to$\,69.67\%). The
improvement grows with temporal separation, indicating that denser mid-tail
sampling improves sensitivity to gradual distributional drift precisely where
prediction is harder. Enabling additive attention in the decision stage lifts
the four-quantile extractor further to 75.92\% and 72.44\% F1 at 30 and 60
minutes, respectively; this attention-enabled four-quantile configuration (P5 with additive attention) initialises the extended-hour and day-scale experiments.

\subsection{Comparison with Baseline Models}

Table~\ref{tab:baselines} benchmarks the proposed TQRNN configurations against
six baselines under the matched short-horizon comparison protocol. The
baseline models are evaluated using the same data representation as the
integrated two-quantile extractor (P4), so that baseline differences primarily reflect modelling capacity rather than variation in feature preprocessing.
The final proposed row reports the integrated four-quantile extractor (P5) with additive attention enabled, which is the configuration used as the
short-horizon bridge for the subsequent extended-horizon experiments.

At the 10-second horizon the task is close to saturation and the competitive
models are tightly clustered: LS-SVM reaches 98.18\% F1, LSRN 98.14\%, and the
stand-alone Transformer 97.25\%, while KNN and SVR trail at 97.32\% and 95.11\%
and LSTM falls back to 87.09\%. The integrated two-quantile extractor (P4) records 97.54\% F1, remaining within half a point of the strongest baselines while
retaining the quantile-integrated feature representation. The integrated four-quantile extractor (P5) with additive attention records the highest value at this horizon, 98.39\% F1, although the margin over the leading baselines is
small because performance is already near ceiling.

As the horizon extends, baseline performance declines more sharply and the
configurations separate. At 30 minutes, SVR falls to 67.02\% F1 and KNN to
61.78\%, while LSTM yields 63.69\% F1. LSRN and LS-SVM provide moderate
resilience at 73.33\% and 69.71\%, respectively, and the stand-alone
Transformer reaches 70.65\%. The integrated two-quantile extractor (P4) records 72.73\% F1, the strongest of the matched configurations excluding the
attention-enhanced bridge, and at 60 minutes records 67.68\% F1, again ahead
of every baseline (best baseline LS-SVM at 66.14\%).

The refined four-quantile extractor (P5) increases mid-tail quantile resolution
from two to four quantiles ($\alpha\in\{0.25,0.40,0.60,0.75\}$) and is
reported here with additive attention enabled. This produces the strongest
F1-score at every short-horizon window. At 30 minutes it achieves 75.92\% F1,
exceeding the two-quantile extractor by 3.19 percentage points. At 60 minutes
it reaches 72.44\% F1 compared with 67.68\%, a gain of 4.76 percentage
points. At the 10-second horizon the underlying two- and four-quantile
extractors are effectively level (97.54\% versus 97.75\% F1,
Table~\ref{tab:short}), as expected where performance is already saturated,
with additive attention contributing the remaining lift to 98.39\%. The
four-quantile representation therefore provides substantially greater
robustness as temporal separation increases, which justifies selecting it,
with additive attention, as the short-horizon bridge configuration for the
70-hour and 30-day experiments.

\begin{table}[!t]
\centering
\caption{Short-Horizon Comparison Against Baseline Models Under the Matched P4 Data Representation. Proposed P5 Reports the Four-Quantile Configuration with Additive Attention Enabled.}
\label{tab:baselines}
\setlength{\tabcolsep}{4.4pt}
\footnotesize
\begin{tabular}{l|cccc|cccc|cccc}
\toprule
& \multicolumn{4}{c|}{10 seconds} & \multicolumn{4}{c|}{30 minutes} & \multicolumn{4}{c}{60 minutes} \\
\cmidrule(lr){2-5}\cmidrule(lr){6-9}\cmidrule(l){10-13}
Model & F1 & Rec & Prec & Acc\% & F1 & Rec & Prec & Acc\% & F1 & Rec & Prec & Acc\% \\
\midrule
SVR & 95.11 & 96.08 & 96.74 & 95.85 & 67.02 & 66.09 & 68.03 & 68.88 & 62.04 & 59.83 & 61.89 & 62.33 \\
KNN & 97.32 & 98.21 & 98.81 & 97.78
    & 61.78 & 60.95 & 63.14 & 61.88
    & 59.67 & 58.36 & 60.58 & 60.71 \\
LSRN & 98.14 & 98.67 & 97.08 & 97.38
    & 73.33 & 71.19 & 73.82 & 73.49
    & 65.11 & 64.83 & 67.08 & 68.17 \\
LS-SVM & 98.18 & 97.54 & 98.02 & 97.18
    & 69.71 & 67.82 & 69.94 & 70.29
    & 66.14 & 65.37 & 66.67 & 68.04 \\
LSTM & 87.09 & 84.67 & 89.92 & 85.39 & 63.69 & 64.16 & 65.28 & 62.93 & 54.66 & 52.18 & 55.72 & 56.91 \\
Transformer & 97.25 & 97.07 & 97.49 & 97.25 & 70.65 & 68.91 & 71.49 & 72.17 & 63.23 & 62.60 & 63.89 & 66.71 \\
Proposed P4: $All^{(QR^{2}_{2Q})}$ & 97.54 & 97.48 & 97.44 & 97.52
& 72.73 & 73.32 & 73.76 & 72.56
& 67.68 & 67.01 & 68.38 & 68.67 \\
Proposed P5 + attn.: $All^{(QR^{2}_{4Q})}$ & \textbf{98.39} & \textbf{98.58} & \textbf{97.92} & \textbf{97.90}
    & \textbf{75.92} & \textbf{75.77} & \textbf{74.33} & \textbf{74.15}
    & \textbf{72.44} & \textbf{72.20} & \textbf{73.65} & \textbf{73.84} \\
\bottomrule
\end{tabular}
\end{table}

\subsection{Extended-Hour Ablation (P5--P9)}

Table~\ref{tab:exthour} evaluates the 36-, 48-, and 70-hour horizons. The
unmodified short-horizon extractor (P5), transferred without adaptation, degrades rapidly, approaching
near-chance performance at 70 hours (42.90\% F1). Increased quantile
resolution alone is therefore insufficient for extended-hour forecasting
unless the architecture is also adapted to the longer temporal regime. The
isolated extended-hour quantile stages (P6, P7), broad and mid-tail alike, outperform the transferred reference but remain limited when used independently, and the
extended-hour Transformer-only baseline (P8) benefits from temporal abstraction but lacks explicit distributional modelling. The fully integrated extended-hour extractor (P9) is strongest across all three horizons, reaching
70.84\%, 66.34\%, and 60.38\% F1 at 36, 48, and 70 hours. Reliable
extended-hour forecasting therefore requires both coarse and refined quantile
structure together with increased feature capacity, and short-horizon
extractors cannot be transferred directly without horizon-appropriate scaling.

\begin{table}[!t]
\centering
\caption{Extended-Hour Feature-Extraction Ablation (Pipelines P5--P9). P5 Is the Unmodified Short-Horizon Reference, Transferred Without Extended-Hour Adaptation.}
\label{tab:exthour}
\setlength{\tabcolsep}{4.4pt}
\footnotesize
\begin{tabular}{l|cccc|cccc|cccc}
\toprule
& \multicolumn{4}{c|}{36 hours} & \multicolumn{4}{c|}{48 hours} & \multicolumn{4}{c}{70 hours} \\
\cmidrule(lr){2-5}\cmidrule(lr){6-9}\cmidrule(l){10-13}
Pipeline & F1 & Rec & Prec & Acc\% & F1 & Rec & Prec & Acc\% & F1 & Rec & Prec & Acc\% \\
\midrule
P5: $All^{(QR^{2}_{4Q})}$ (transferred) & 48.76 & 49.11 & 49.99 & 50.61 & 44.70 & 45.46 & 45.66 & 45.78 & 42.90 & 42.19 & 43.93 & 44.33 \\
P6: $QR^{1}_{70h}$ & 53.92 & 54.27 & 55.15 & 55.77 & 52.86 & 51.62 & 52.82 & 51.94 & 50.06 & 51.35 & 51.09 & 50.49 \\
P7: $QR^{2}_{70h}$ & 57.27 & 58.06 & 58.75 & 58.81 & 54.21 & 54.96 & 55.31 & 55.13 & 51.76 & 52.34 & 52.83 & 52.77 \\
P8: Transformer$_{70h}$ & 67.78 & 67.97 & 68.06 & 67.12 & 61.87 & 60.56 & 61.61 & 61.09 & 56.63 & 57.03 & 56.96 & 57.09 \\
P9: $All_{70h}$ & \textbf{70.84} & \textbf{70.37} & \textbf{71.74} & \textbf{72.06} & \textbf{66.34} & \textbf{65.79} & \textbf{65.20} & \textbf{66.10} & \textbf{60.38} & \textbf{61.87} & \textbf{61.40} & \textbf{61.89} \\
\bottomrule
\end{tabular}
\end{table}

\subsection{Day-Scale Ablation (P9--P13)}

Table~\ref{tab:thirty} reports the 7-, 14-, and 30-day results, evaluating in
particular the impact of expanding the sensor-level input dimensionality from
43 to 81 channels. The integrated extended-hour extractor (P9) is retained as the transferred bridge configuration and is evaluated directly at the day-scale
horizons before the full \TQRNNd{} feature extensions are introduced. Its
performance declines
from 61.49\% F1 at 7 days to 52.41\% F1 at 30 days, confirming that the
strongest extended-hour configuration does not generalise directly to
long-term degradation modelling without day-scale adaptation.

\begin{figure}[!t]
\centering
\begin{tikzpicture}
\begin{axis}[
width=0.98\linewidth, height=58mm,
xlabel={Forecast horizon (days)}, ylabel={F1-score (\%)},
xtick={7,14,30}, ymin=45, ymax=85,
legend style={font=\tiny, at={(0.5,-0.32)}, anchor=north, legend columns=3,
/tikz/every even column/.append style={column sep=4pt}},
grid=both, grid style={black!10},
tick label style={font=\scriptsize}, label style={font=\scriptsize},
]
\addplot[black!45, thin, mark=o, mark size=1.4pt]
coordinates {(7,61.49) (14,57.58) (30,52.41)};
\addlegendentry{P9 $All_{70h}$ bridge}
\addplot[black!35, thin, mark=square, mark size=1.3pt]
coordinates {(7,56.59) (14,52.44) (30,50.60)};
\addlegendentry{P10 $QR^{1}_{30d}$}
\addplot[black!55, thin, mark=diamond, mark size=1.6pt]
coordinates {(7,61.81) (14,58.68) (30,56.17)};
\addlegendentry{P11 $QR^{2}_{30d}$}
\addplot[name path=ptwelve, orange!85!black, thick, mark=triangle*, mark size=1.8pt]
coordinates {(7,72.56) (14,66.52) (30,67.13)};
\addlegendentry{P12 Transformer$_{30d}$}
\addplot[name path=pthirteen, blue!70!black, very thick, mark=*, mark size=1.8pt]
coordinates {(7,76.09) (14,76.76) (30,79.97)};
\addlegendentry{P13 $All_{30d}$}
\addplot[blue!9] fill between[of=pthirteen and ptwelve];
\node[font=\tiny, blue!60!black] at (axis cs:8.4,74.9) {+3.5};
\node[font=\tiny, blue!60!black] at (axis cs:14,72.2) {+10.2};
\node[font=\tiny, blue!60!black, anchor=east] at (axis cs:29.2,74.0) {+12.8};
\node[font=\tiny, align=left, anchor=west, text=black!70]
at (axis cs:16.5,82.3) {only P13 improves with horizon};
\end{axis}
\end{tikzpicture}
\caption{F1-score versus day-scale horizon. Only the fully integrated
horizon-conditioned extractor (P13) improves with horizon. The shaded band
marks its margin over the strongest alternative configuration
(P12, Transformer$_{30d}$), which widens from $+3.5$ to $+12.8$ F1 points
between the 7- and 30-day horizons; the transferred bridge (P9) and
isolated quantile stages (P10, P11) decline throughout.}
\label{fig:dayplot}
\end{figure}
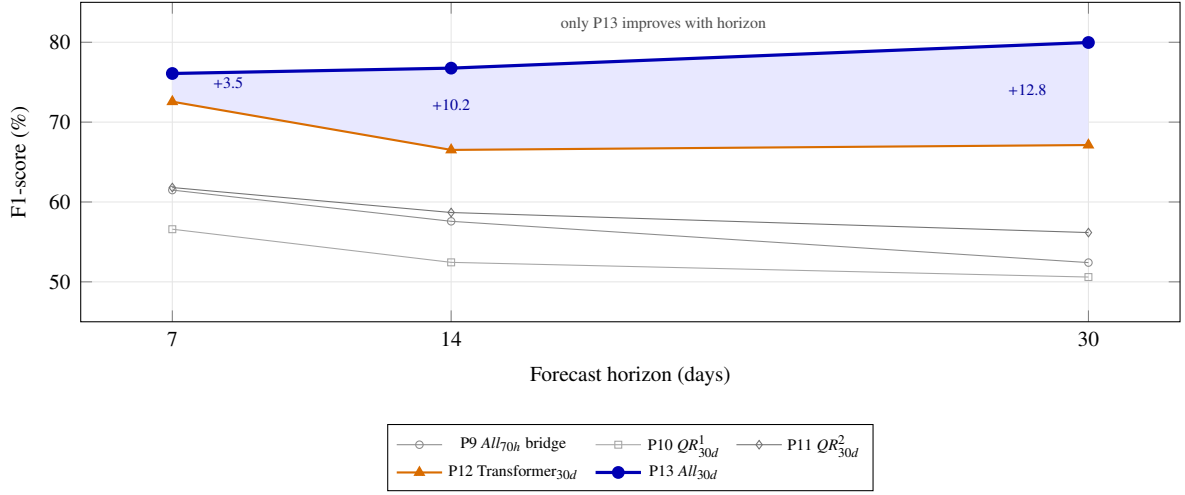

The isolated day-scale quantile stages (P10, P11) add capacity but remain constrained when used without full hierarchical integration. The day-scale Transformer-only variant (P12) benefits from greater temporal abstraction and
outperforms the isolated quantile stages, but still underperforms the fully
integrated quantile-led configuration. The complete integrated day-scale extractor (P13) is strongest at every day-scale horizon and, notably,
\emph{improves} with horizon, from 76.09\% F1 at 7 days to 79.97\% F1,
80.18\% recall, 81.82\% precision, and 82.39\% accuracy at 30 days, while
the transferred, isolated, and Transformer-only configurations decline
(Fig.~\ref{fig:dayplot}). Stable long-horizon prediction therefore requires
richer quantile representations, increased feature-extraction depth,
expanded sensor breadth, and hierarchical integration capable of capturing
slow-evolving degradation patterns and cumulative distributional shift.

\begin{table}[!t]
\centering
\caption{Day-Scale Feature-Extraction Ablation (Pipelines P9--P13), Evaluating the 43-to-81-Channel Input Expansion. P9 Is the Transferred \TQRNNh{} Bridge Configuration Evaluated at Day Scale.}
\label{tab:thirty}
\setlength{\tabcolsep}{4.4pt}
\footnotesize
\begin{tabular}{l|cccc|cccc|cccc}
\toprule
& \multicolumn{4}{c|}{7-day} & \multicolumn{4}{c|}{14-day} & \multicolumn{4}{c}{30-day} \\
\cmidrule(lr){2-5}\cmidrule(lr){6-9}\cmidrule(l){10-13}
Pipeline & F1 & Rec & Prec & Acc\% & F1 & Rec & Prec & Acc\% & F1 & Rec & Prec & Acc\% \\
\midrule
P9: $All_{70h}$ transferred bridge & 61.49 & 61.08 & 62.27 & 62.55
& 57.58 & 57.11 & 56.59 & 57.37
& 52.41 & 53.70 & 53.30 & 53.72 \\
P10: $QR^{1}_{30d}$ & 56.59 & 56.95 & 57.85 & 58.48
 & 52.44 & 53.21 & 53.42 & 53.54
 & 50.60 & 49.87 & 51.65 & 52.06 \\
P11: $QR^{2}_{30d}$ & 61.81 & 62.62 & 63.33 & 63.39
& 58.68 & 59.45 & 59.81 & 59.62
& 56.17 & 56.77 & 57.27 & 57.21 \\
P12: Transformer$_{30d}$ & 72.56 & 72.81 & 72.92 & 71.93
& 66.52 & 65.19 & 66.26 & 65.76
& 67.13 & 66.54 & 66.47 & 67.51 \\
P13: $All_{30d}$ & \textbf{76.09} & \textbf{75.81} & \textbf{76.89} & \textbf{77.14}
 & \textbf{76.76} & \textbf{78.73} & \textbf{78.31} & \textbf{78.69}
 & \textbf{79.97} & \textbf{80.18} & \textbf{81.82} & \textbf{82.39} \\
\bottomrule
\end{tabular}
\end{table}

\subsection{Cross-Regime Interpretation}

Three findings hold across the thirteen pipelines. First,
\emph{quantile integration matters at every horizon}: the Transformer-only configurations (P3, P8, P12) are competitive in all three regimes but never strongest,
confirming that explicit distribution-aware feature extraction adds
information that temporal modelling alone does not recover. Second,
\emph{mid-tail resolution is a robust representational choice}: the
four-quantile set introduced at P5 is retained by every
subsequent winning configuration, and its benefit grows with temporal
separation. Third, and most importantly, \emph{the benefit is conditional on
horizon-appropriate scaling}: the progression P5\,$\to$\,P9\,$\to$\,P13 is cumulative, while direct transfer of a representation beyond its design limits produces severe degradation. P5 collapses from 72.44\% F1 at 60 minutes to 42.90\% at 70 hours when the short-horizon extractor is transferred without extended-hour adaptation. Similarly, the P9 bridge declines from its 70-hour design-point result of
60.38\% F1 to 52.41\% F1 at 30 days when transferred to day-scale prediction
without the full \TQRNNd{} feature extensions. In contrast, P13 reaches 79.97\% F1 at 30 days after the 81-channel sensor expansion, increased
feature capacity, temporal embedding, and full day-scale quantile integration
are introduced. In representational terms, day-scale quantile surfaces
display substantially richer differentiation in the upper quantile bands than
the corresponding 1-hour representations, indicating that deeper temporal
context improves the legibility of health trajectories rather than merely
increasing model complexity. Feature extraction in PdM is therefore best
treated as a horizon-dependent design choice, with quantile cardinality,
capacity, temporal embedding, activation, and sensor breadth co-designed with
the forecasting task.

\subsection{Scope and Limitations}

The evaluation is deployment-scale but not sector-universal. All machines
belong to the same broad manufacturing asset and product family, and the
reported gains are conditioned on the available sensor ecology, operating
regimes, and labelling process. The staged transfer experiments therefore
support strong claims for this industrial context, but they should not be
read as evidence that the same absolute performance levels will transfer
unchanged to different sectors, asset classes, or maintenance cultures. The
two-tier training procedure freezes QRNN$_1$ before refinement, which
stabilises optimisation but may forgo potential gains from joint fine-tuning.
The retained quantile levels were selected by validation within the evaluated
asset family rather than by exhaustive search over all quantile combinations.
Ablation results are reported as point estimates under matched partitions;
fold-level paired significance testing of the final long-horizon
configuration against external baselines is reported in the companion
long-horizon study.
Further validation is therefore required before generalising the same
horizon-conditioned scaling principle to substantially different industrial
systems.

\section{Conclusion}
\label{sec:conclusion}

This paper presented a quantile-led feature-extraction framework for
multi-horizon predictive maintenance, in which a dual-stage MLP--QRNN
hierarchy converts high-volume multivariate sensor streams into compact,
channel-resolved, distribution-aware feature vectors. Under channel-resolved
pinball-loss supervision, a broad ten-quantile first stage and a
skip-connected mid-tail refinement stage compresses roughly 180{,}000
reference-grid samples per channel each hour into feature vectors of dimension
$d_f\in\{86,172,324\}$. A fixed thirteen-pipeline staged ablation across
1-hour, 70-hour, and 30-day regimes on 72 machines in 9 facilities showed
that four-quantile mid-tail refinement
($\alpha\in\{0.25,0.40,0.60,0.75\}$) improves robustness as temporal
separation increases. The results also show that the representation transfers
across horizons only when feature capacity, temporal embedding, activation
strategy, and sensor breadth are scaled to the forecasting task. The
horizon-conditioned extractors reach 60.38\% F1 at 70 hours and 79.97\% F1
at 30 days, while the transferred P9 bridge declines to 52.41\% F1 at 30 days without full day-scale adaptation. Quantile-integrated pipelines
outperform Transformer-only and isolated-stage alternatives at every horizon.
The central conclusion is therefore that feature extraction for industrial
prognostics should be designed as a horizon-dependent representational stage
rather than fixed preprocessing. Future work will examine joint end-to-end
fine-tuning of the two quantile stages, learned selection of retained
quantile levels, and cross-sector transfer of the horizon-conditioned scaling
principle.

\section*{Acknowledgment}
The authors thank the partner manufacturing organisation for access to
deployment data, operational context, and engineering support.

\section*{Declaration of competing interest}
The authors declare that they have no known competing financial interests
or personal relationships that could have appeared to influence the work
reported in this paper.

\section*{Data availability}

The industrial sensor and maintenance data used in this study are
proprietary to the partner manufacturing organisation and are subject to a
confidentiality agreement; they are therefore not publicly available. Data
may be made available by the corresponding author upon reasonable request
and with the permission of the partner organisation.

\end{document}